\documentclass[10pt,twocolumn,letterpaper]{article}

\usepackage[pagenumbers]{wacv} % To force page numbers, e.g. for an arXiv version

\usepackage{graphicx}
\usepackage{amsmath}
\usepackage{amssymb}
\usepackage{booktabs}
\usepackage{comment}
\usepackage[table]{xcolor}
\usepackage[pagebackref,breaklinks,colorlinks]{hyperref}

\usepackage[capitalize]{cleveref}
\crefname{section}{Sec.}{Secs.}
\Crefname{section}{Section}{Sections}
\Crefname{table}{Table}{Tables}
\crefname{table}{Tab.}{Tabs.}

\def\wacvPaperID{2462} % *** Enter the WACV Paper ID here
\def\confName{WACV}
\def\confYear{2025}

\begin{document}

%\title{Effective Initial Convolution Enables Scalable, Tokenization-Free Hardware-Friendly Diffusion Models Architectures with Fixed-Size Reusable Structures for On-Device Image Generation}

% \title{Scalable, Tokenization-Free Diffusion Model Architectures with Efficient Initial Convolution and Fixed-Size Reusable Structures for On-Device Image Generation}

%\title{Scalable, Tokenization-Free Diffusion Model Architectures with Fixed-Size Reusable Structures for On-Device Image Generation}

%\title{Optimizing Initial Convolution for Scalable, Token-Free Vision Transformer Architectures in Hardware-Efficient Diffusion Models}

%\title{Enhanced Initial Convolution for Scalable and Tokenization-Free Vision Transformer Architectures in Hardware-Optimized Diffusion Models}

\title{Hardware-Friendly Diffusion Models with Fixed-Size Reusable Structures for On-Device Image Generation}

%\title{Streamlining Vision Transformers: Effective Initial Convolution for Scalable, Token-Free Architectures in Diffusion Models}

%\title{Scalable Vision Transformer Architectures for Diffusion Models: Leveraging Initial Convolution for Token-Free and Hardware-Friendly Designs}

\author{Sanchar Palit\\
IIT Bombay\\
\and
Sathya Veera Reddy Dendi\\
Samsung\\
\and
Mallikarjuna Talluri \\
Samsung\\ 
\and
Raj Narayana Gadde\\
Samsung
}
\maketitle

%%%%%%%%% ABSTRACT
\begin{abstract}
  Vision Transformers and U-Net architectures have been widely adopted in the implementation of Diffusion Models. However, each architecture presents
  specific challenges while realizing them on-device. Vision Transformers require positional embedding to maintain correspondence between the tokens processed by the transformer, although they offer the advantage of using fixed-size, reusable repetitive blocks following tokenization. The U-Net architecture lacks these attributes, as it utilizes variable-sized intermediate blocks for down-convolution and up-convolution in the noise estimation backbone for the diffusion process. To address these issues, we propose an architecture that utilizes a fixed-size, reusable transformer block as a core structure, making it more suitable for hardware implementation. Our architecture is characterized by low complexity, token-free design, absence of positional embeddings, uniformity, and scalability, making it highly suitable for deployment on mobile and resource-constrained devices. The proposed model exhibit competitive and consistent performance across both unconditional and conditional image generation tasks. The model achieved a state-of-the-art FID score of 1.6 on unconditional image generation with the CelebA. %and a competitive FID score of -- on the MSCOCO for text-conditional image generation.

\end{abstract}

%%%%%%%%% BODY TEXT

\section{Introduction}
\label{sec:intro}

\begin{figure}[t]
  \centering
   \includegraphics[width=1.0\linewidth]
    {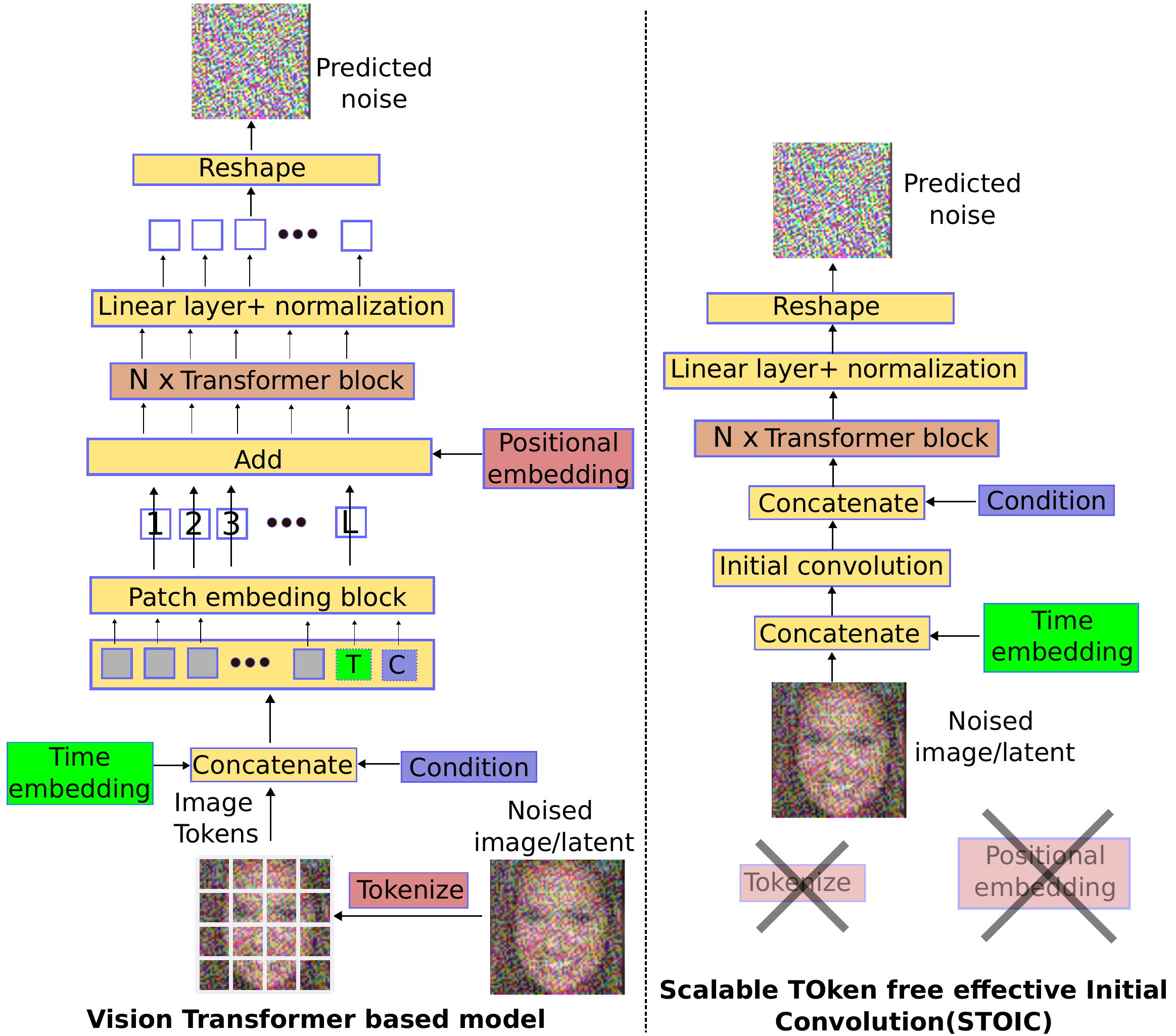}
   %{Fig1.pdf}

   %\caption{Our model eliminates the complex requirements of tokenization and positional embedding present in Vision Transformers. This simplification makes our model more suitable for hardware implementation, as it utilizes a single, repetitive core block enhancing hardware efficiency and feasibility. Additionally, in our architecture, the core blocks can be scaled both by increasing the number of blocks and by adjusting the embedding dimension.}
   \caption{Comparative neural network architectures, contrasting the Vision Transformer (ViT) on the left with the proposed hardware-friendly, low-complexity, scalable, tokenization-free neural network on the right, which features positional embeddings free transformer block-based model.}
   \label{fig:unet_vit_core}
\end{figure}

\begin{figure*}
    \centering
    \includegraphics[width=1.0\linewidth]
    {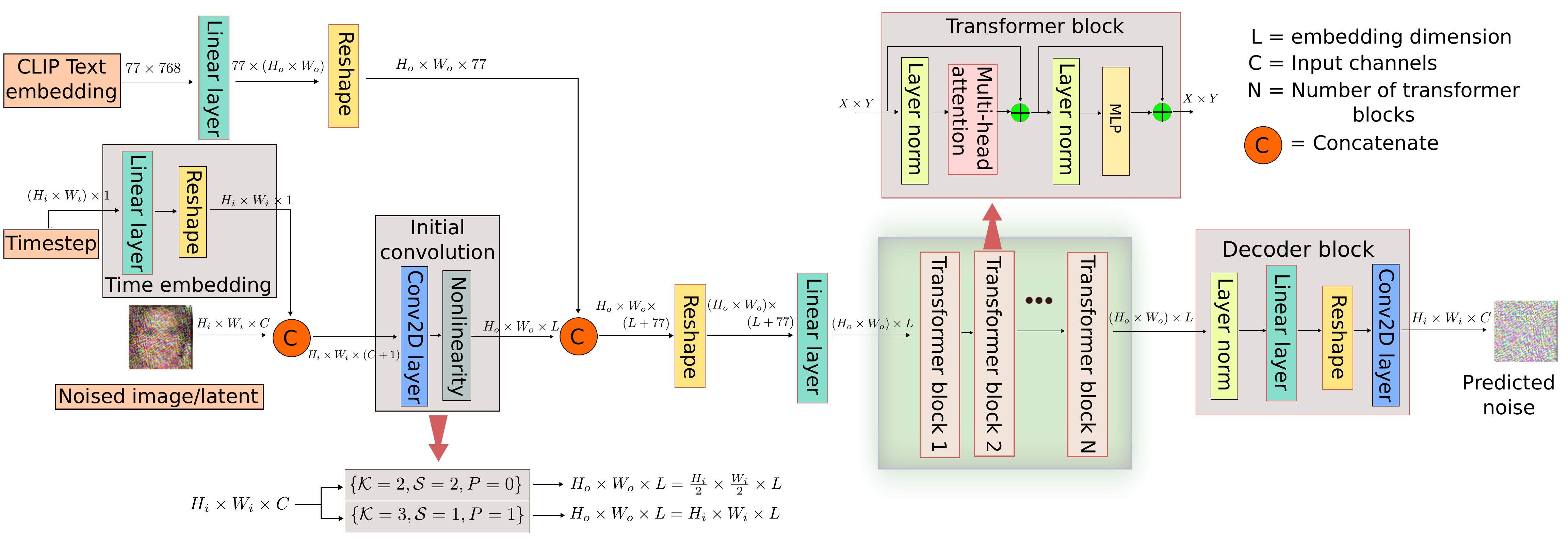}
    %{figures/T2I_mscoco_core_lite.pdf}
    
     \caption{Proposed hardware-friendly architecture of the text-conditional diffusion model in the image space or the latent space. The core structure (transformer block) produces an output with the same dimensions as the input. The input and output dimensions of the core structure are determined based on the initial convolution. For unconditional image generation, the output of the initial convolution is directly passed to the core structures for further processing. Text embeddings are derived from the CLIP~\cite{radford2021learning} text encoder, while image embeddings are obtained using the Latent Diffusion Model~\cite{rombach2022high} encoder.}
     \label{fig:core_structure}
\end{figure*}

Diffusion Models~\cite{sohl2015deep,ho2020denoising,song2020score} have gained popularity in recent generative model tasks due to their stable training compared to Generative Adversarial Networks~\cite{goodfellow2020generative,xu2018attngan, zhu2019dm, gu2022vector, zhang2021cross} and their effective image 
sampling using robust strategies such as classifier-free guidance~\cite{ho2022classifier}. Consequently, Diffusion Models are widely employed in various image generation tasks~\cite{choi2021ilvr,zhao2022egsde, saharia2022palette}, including text-to-image 
generation~\cite{ramesh2021zero,ramesh2022hierarchical,ding2021cogview,rombach2022high,nichol2021glide,saharia2022photorealistic}, image super-resolution~\cite{gao2023implicit,yue2024resshift}, and layout-to-image generation~\cite{zheng2023layoutdiffusion,couairon2023zero}. These models operate through a forward noising process and a backward denoising process, both implemented using 
Markov Chains~\cite{norris1998markov}. The backward or reverse processes are typically realized using CNN-based architectures, such as U-Net~\cite{ronneberger2015u}, or Vision Transformer-based models~\cite{peebles2023scalable,bao2023all,yang2022your}. U-Net-based models~\cite{ho2020denoising, song2020score, rombach2022high} have proven to be highly effective in implementing the diffusion mechanism. These models efficiently capture the spatial structure of images through a sequence of downsampling, mid-layer convolution, and upsampling stages. This makes the U-Net structure challenging to implement on devices, where varying block sizes are required to handle differently in terms of resource allocation on hardware.

In contrast, Vision Transformers~\cite{khan2022transformers} exhibit scalability while maintaining consistent intermediate block sizes; however, they introduce the overhead of tokenization and additional positional embeddings to ensure coherence among the tokens derived from the input image, which results in increased model latency. Consequently, training and inference times would increase, making it difficult to deploy on real-time systems.

This raises the question: \textit{Can we design an architecture with the same-sized repetitive blocks as transformer blocks and without the additional overhead of positional embeddings and tokens?}

This work introduces a hardware-friendly neural network architecture for image generation using a diffusion model. This architecture is distinguished by its low complexity, token-free design, uniformity, scalability, and hardware efficiency. The core component of our architecture is the initial convolution blocks, which serve as the basis for two different configurations. In Configuration I (stride $\mathcal{S}=2$), increasing the parameter count by expanding the embedding dimension or the number of reusable core structures (transformer blocks) results in a minimal increase in computational complexity while preserving high-quality image generation. In contrast, in Configuration II (stride $\mathcal{S}=1$), increasing the parameter count by similarly expanding the embedding dimension or the number of reusable core structures leads to higher computational complexity. However, the performance of Configuration II is superior when compared to Configuration I.
 
 Following the initial convolution, the entire image is processed through uniform-sized core structures, similar to those in Vision Transformers, which incorporate attention mechanisms and MLP layers. Additionally, the initial convolution block in our architecture effectively eliminates the overhead associated with tokenization and removes the need for positional embeddings, as required in Vision Transformers (as illustrated in Figure~\ref{fig:unet_vit_core}). These core structures can be scaled to achieve higher image quality, as measured by the FID~\cite{heusel2017gans} score. The initial convolution is crucial for extracting comprehensive spatial information from the image and combines the time embedding information with feature maps of the image, while the subsequent transformer blocks are responsible for extracting meaningful information from the features during the denoising phase of the diffusion process.

%\textcolor{red}{Next, we provide the background for our work, followed by a discussion of the methodology and implementation details. We then review related work and conclude with our experimental results and analysis.}
%-------------------------------------------------------------------------
\begin{figure*}
\centering    
\begin{subfigure}{0.32\linewidth}
  \centering
   \includegraphics[width=1.0\linewidth]{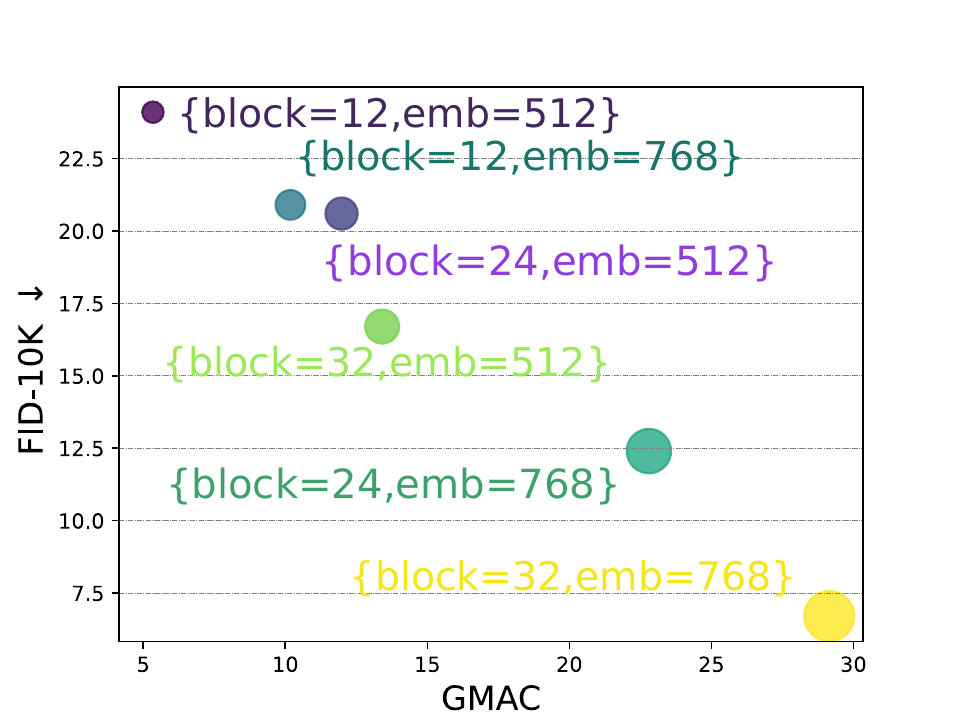}
   \caption{}
   %\caption{GFlops with different embedding dimension and different number of blocks for $\mathcal{S}=2$.}
   \label{fig:gflops_s1_s2}
\end{subfigure}
\begin{subfigure}{0.32\linewidth}
  \centering
   \includegraphics[width=1.0\linewidth]{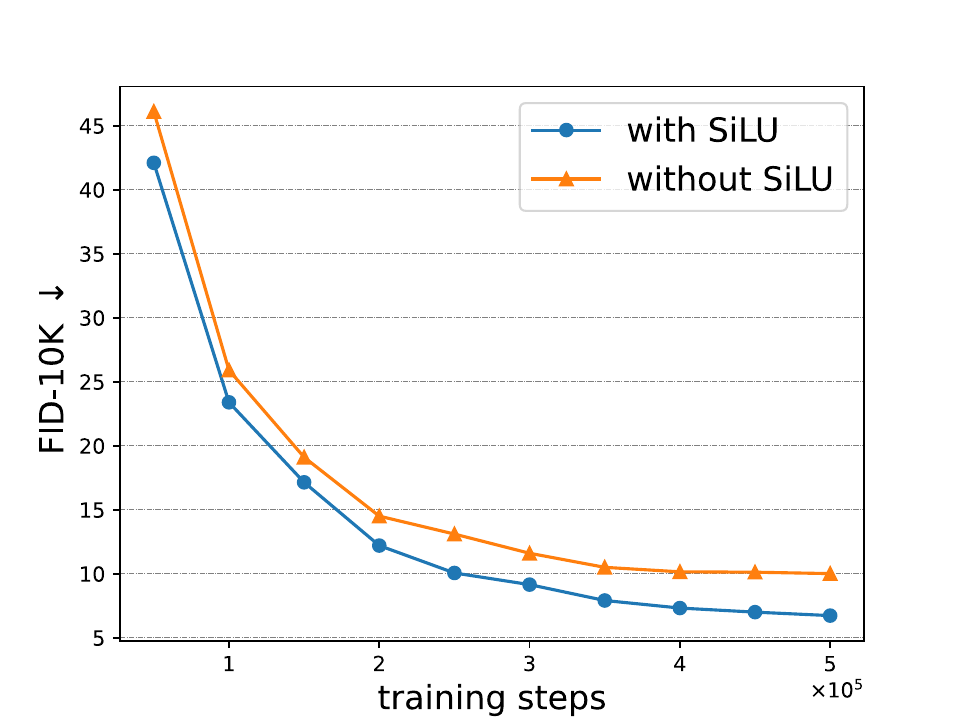}
   \caption{}
   %\caption{Non linearity in initial embedding layer}
   \label{fig:nonli_ini}
\end{subfigure}
\begin{subfigure}{0.32\linewidth}
  \centering
   \includegraphics[width=1.0\linewidth]{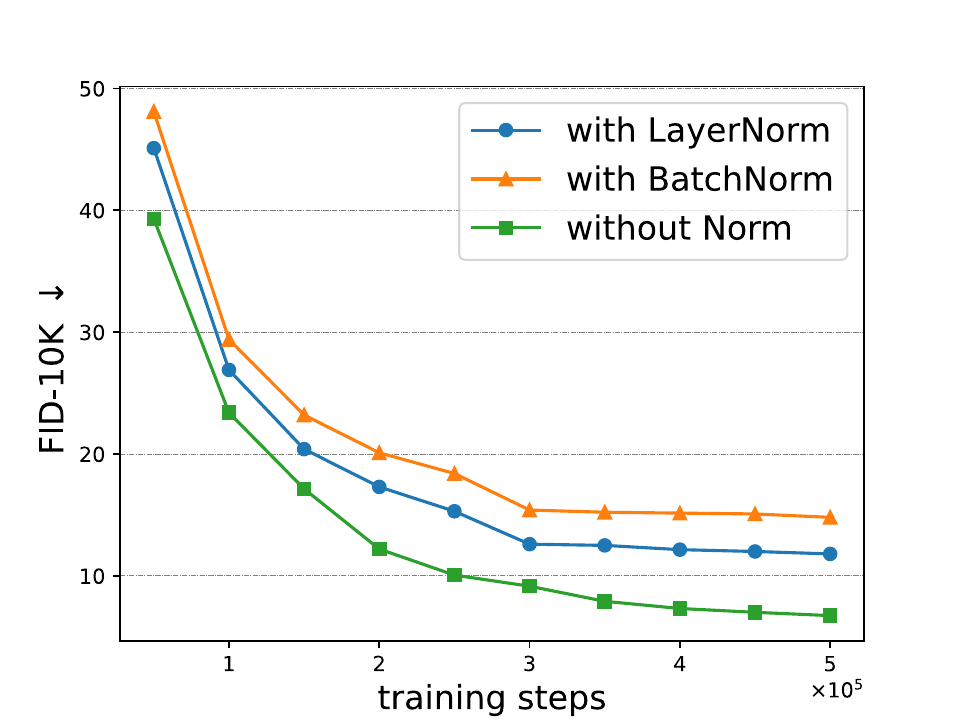}
   \caption{}
   %\caption{Normalization in initial embedding layer}
   \label{fig:norm_ini}
\end{subfigure}
\begin{subfigure}{0.32\linewidth}
  \centering
   \includegraphics[width=1.0\linewidth]{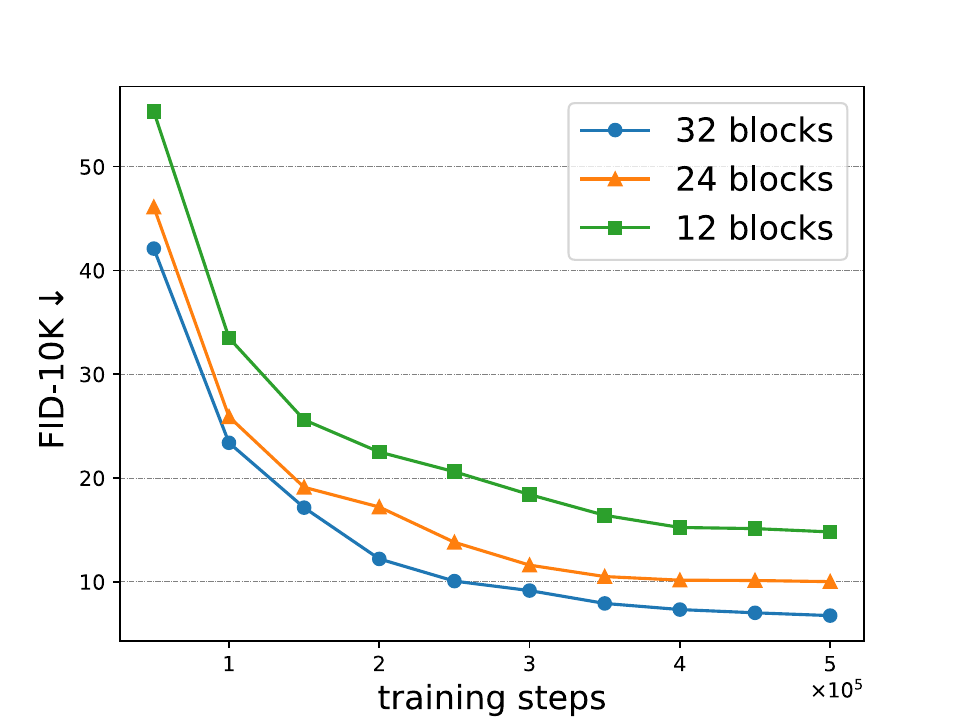}
   \caption{}
   %\caption{Different number of Core blocks}
   \label{fig:mod_si}
\end{subfigure}
\begin{subfigure}{0.32\linewidth}
  \centering
   \includegraphics[width=1.0\linewidth]{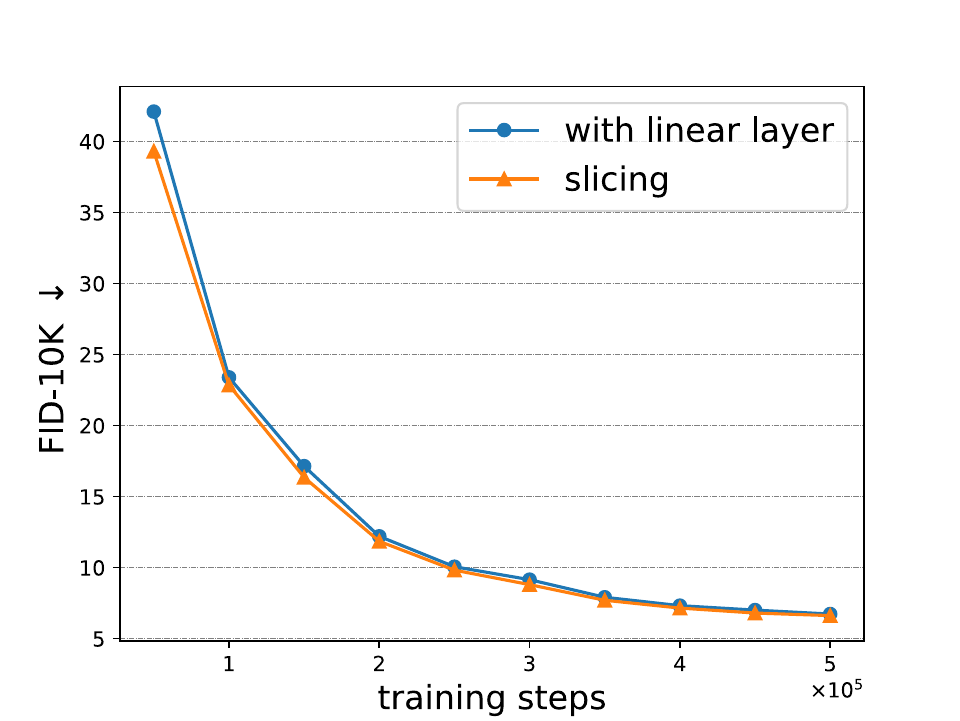}
   \caption{}
   %\caption{Linear layer vs slicing operation in the Decoder block.}
   \label{fig:lin_sli}
\end{subfigure}
\begin{subfigure}{0.32\linewidth}
  \centering
   \includegraphics[width=1.0\linewidth]{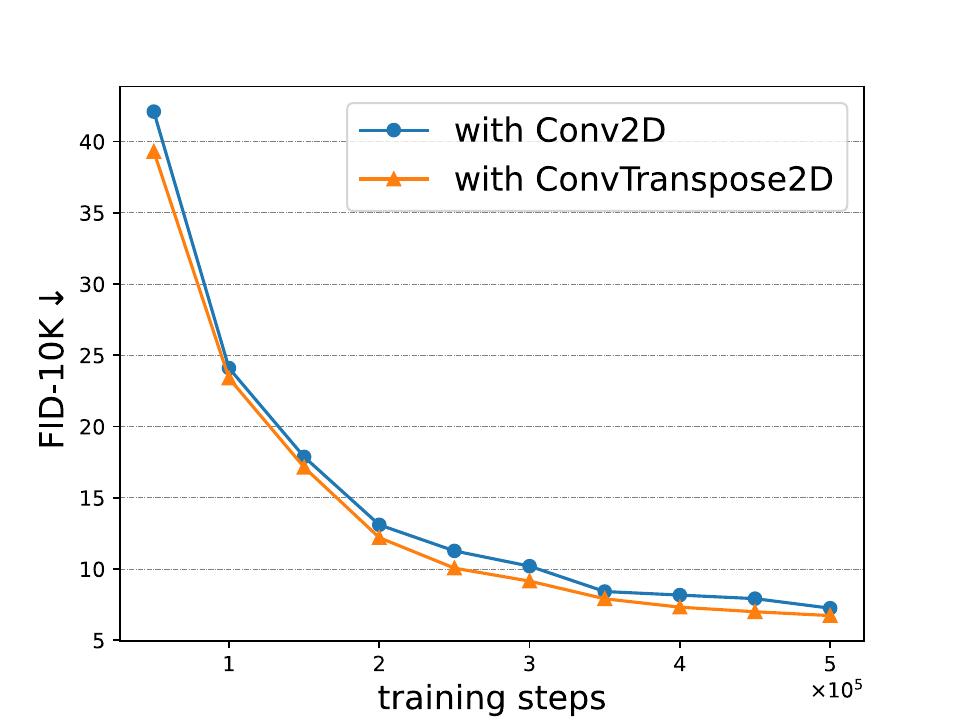}
   \caption{}
   %\caption{ Conv2D vs ConvTranspose2D operation in the decoder block}
   \label{fig:conv_conv_t}
\end{subfigure}
\caption{Ablation study of design choices of the proposed architecture on CIFAR-10. (a) Comparison of FID vs GMAC with different embedding dimensions and number of blocks with configuration-I (stride $\mathcal{S}=2$). (b) Non-linearity in the initial convolution block. (c) Normalization in the initial convolution block. (d) Number of transformer blocks (12, 24 and  32). (e) Linear layer vs slicing operation in the Decoder block. (f) Conv2D vs ConvTranspose2D operation in the decoder block}
\end{figure*}

\section{Background}
\subsection{Diffusion Models}
Diffusion models such as Denoising Diffusion Probabilistic Models (DDPM)~\cite{ho2020denoising} train a model to generate images from standard Gaussian noise. Consequently, the model operates in two phases, each formulated as a Markov chain: the forward process and the reverse process.
%where noise is incrementally added to the image at successive timesteps, and the backward process, where noise is gradually removed at each timestep to reconstruct the original image from Gaussian noise $\mathcal{N}(0,\mathrm{I})$.
The forward process, which incrementally adds noise to the image at successive timesteps, is defined as follows:
$$ \mathcal{Q}(x_T|x_0) = \Pi_{t=1}^{T}\mathcal{Q}(x_t|x_{t-1}) = \mathcal{N}(\sqrt{\Bar{\alpha}_T}x_0,(1-\Bar{\alpha}_T)\mathrm{I}) 
$$ where the intermediate marginals are given by
$\mathcal{Q}(x_t|x_{t-1}) := \mathcal{N}(x_t|\alpha_t x, {\sigma_t}^2\mathrm{I})$.
Here, $\alpha_t, \sigma_t \in (0,1)$ denote the shift and change in uncertainty, respectively, from the previous data point due to the addition of noise and $\Bar{\alpha}_T = \Pi_{t=1}^T\alpha_t$. The ratio $\frac{\alpha_t^2}{\sigma_t^2}$
 represents the signal-to-noise ratio (SNR) after noise injection, which decreases monotonically. We assume a variance-preserving forward process~\cite{kingma2021variational}, i.e., $\alpha_t = \sqrt{(1-\sigma_t^2)}$. During the reverse, or denoising process, noise is gradually removed at each timestep, and DDPM learns to reconstruct the image from $\mathcal{N}(x_T;0,\mathrm{I})$. The denoising process, $\mathcal{Q}(x_{t-1}|x_t) = \mathcal{N}(x_{t-1}|\mu_{\theta}(x_t,t), \Sigma_{\theta}(x_t,t))$, applies maximum log-likelihood to the data points $x_0$, with $\mu_{\theta}$ and $\Sigma_{\theta}$ parameterized by a neural network model. %Maximizing the data log-likelihood can be accomplished using the Evidence Lower Bound (ELBO) loss in variational inference, expressed as $\mathbb{E}[-\log \mathbb{P}_{\theta}(x)] \leq \mathbb{E}[-\log {\mathbb{P}_{\theta}(x}|\mathcal{Q}(x))]$.
The tractable denoising process $\mathcal{Q}(x_{t-1}|x_t,x_0)$ is then compared against the forward process using KL divergence. This can be simplified to a loss similar to the Mean Squared 
Error (MSE)~\cite{kingma2021variational} on the noise variable $\epsilon_t$. 
%as other losses, such as the prior loss and the reconstruction loss, effectively equate to zero. 
Using the reparameterization of the marginals $\mathcal{Q}(x_{t-1}|x_t,x_0)$, 
the variable  $x_t = \alpha_t x + \sigma_t \epsilon_t$ is predicted in the noise prediction model $\hat{\epsilon}_t$ to obtain the data points $\hat{x}_{\theta}(x_t;t) = x_t/\alpha_t - \sigma_t \hat{\epsilon}_{\theta}(x_t;t)/{\alpha_t}$. Thus, during the reverse process, we aim to learn the injected noise by minimizing the DDPM~\cite{ho2020denoising} loss, expressed as:
\begin{equation}
\underset{\theta}{\arg \min} \mathbb{E} \lvert\lvert \epsilon -\epsilon_{\theta}(x_t, t)\rvert\rvert_2^2
\label{eq:ddpm_loss}
\end{equation}
The diffusion process was subsequently proposed to be implemented using a score network to estimate the noise collectively through various techniques~\cite{song2019generative,song2020score,song2020improved}, such as stochastic differential equations~\cite{song2020score}. These approaches diffuse the data distribution into a fixed prior distribution, enabling sample generation by reversing the diffusion process over time. This method learns the diffusion process by minimizing the Noise Conditional Score Network (NCSN) score loss~\cite{song2020score,song2019generative}.
\begin{equation}
    \mathbb{E}_{x_0,x_t}[\lvert\lvert s_{\theta}(x_t,t) - \nabla_{x_t} \log \mathbb{P} (x_t|x_0)\rvert\rvert_2^2]
    \label{eq:ncsn_1}
\end{equation}
By considering the transition distribution as a Gaussian distribution, $\mathbb{P}(x_t|x_0) = \mathcal{N}(x_t;\mu(t)x_0, \sigma^2(t)\mathrm{I})$ we obtain $\nabla_{x_t} \log \mathbb{P} (x_t|x_0) = \frac{x_t-\mu(t)x_0}{\sigma^2(t)} = \epsilon$. Consequently, Equation.~\ref{eq:ncsn_1} effectively becomes:
\begin{equation}
    \mathbb{E}_{x_0,x_t}[\lvert\lvert \sigma(t) s_{\theta}(\mu(t) x_0 + \sigma_t\epsilon,t) - \epsilon \rvert\rvert_2^2]
    \label{eq:ncsn_2}
\end{equation}
This is equivalent to the DDPM loss when assuming a Gaussian distribution, where $\sigma(t) s_{\theta}(\mu(t) x_0 + \sigma(t) \epsilon,t) = \epsilon_{\theta} (\mu(t) x_0 + \sigma(t) \epsilon,t)$.

\begin{comment}
\begin{figure}[t]
  \centering
   \includegraphics[width=1.0\linewidth]{figures/fid_plt_core_block_num.pdf}
   \caption{Different number of Core blocks}
   \label{fig:mod_si}
\end{figure}

\begin{figure}[t]
  \centering
   \includegraphics[width=1.0\linewidth]{figures/fid_plt_lin_sli.pdf}
   \caption{Linear layer vs slicing operation in the Decoder block.}
   \label{fig:lin_sli}
\end{figure}

\begin{figure}[t]
  \centering
   \includegraphics[width=1.0\linewidth]{figures/fid_plt_conv_convt.pdf}
   \caption{ Conv2D vs ConvTranspose2D operation in the decoder block}
   \label{fig:conv_conv_t}
\end{figure}
\end{comment}

\subsection{Latent Diffusion}
%Diffusion models are highly efficient for image generation. However, training high-resolution images with any neural network-based model using stochastic gradient descent is challenging. 

Diffusion models are highly effective for image generation, but training them for high-resolution outputs presents significant challenges. To overcome this, Latent Diffusion Models (LDM) ~\cite{rombach2022high} have gained popularity, as they carry out the diffusion process in a latent space. The LDM addresses this issue by encoding the image into latent space, $\mathcal{Z}=\mathcal{E}(\mathcal{X})$. The subsequent diffusion process is then applied in the latent space, where $\mathcal{Z} \in \mathbb{R}^{H_i \times W_i}$ has a lower 
resolution than the original image $\mathcal{X}$. Finally, the image is regenerated from the latent space using a decoder, $\Hat{\mathcal{X}}=\mathcal{D}(\mathcal{Z})$.

\subsection{Architectures for Diffusion Models}

%\textcolor{red}{In diffusion models, the forward process (adding noise) is done using different known noise scheduling \cite{kingma2021variational}. The reverse process (denoising), which is needed for generation, is parameterized by a neural network.}

The neural networks used for noise estimation during the time-reversal denoising process are predominantly designed with Vision Transformer (ViT) and U-Net architectures.

Several variants of ViT have been proposed for diffusion models, such as DiT \cite{peebles2023scalable}. Similarly, other architectures include Gen-ViT~\cite{yang2022your}, which does not use any output convolution and integrates the time token before the initial embedding. A closely related work is the U-ViT~\cite{bao2023all} model, which employs Vision Transformers along with long skip connections inspired by the U-Net architecture. U-ViT utilizes positional embeddings, similar to other Vision Transformer-based models, and delivers better results with smaller patch sizes. The disadvantage of long skip connections is that they require storing the initial tensors before finally adding them to the sequential outputs of the transformer blocks. This results in increased memory requirements and latency from a hardware implementation perspective.

The U-Net~\cite{ronneberger2015u}-based model~\cite{rombach2022high,song2020score,ho2020denoising}, popular in image segmentation tasks, employs initial down convolution followed by mid convolution and up convolution to capture the spatial information of images. Various diffusion models, such as Stable Diffusion~\cite{rombach2022high}, have utilized U-Net-based architecture as a backbone. These models typically incorporate attention blocks to condition the image generation task on text embedding. U-ViT too have several challenges while realizing such model on-devices like skip connection, varying feature map dimension and difficult to scale. 

In the following section, we introduce the design of a neural network that combines the strengths of both Vision Transformers (ViT) and U-Net architectures while addressing their respective drawbacks. This is achieved by designing a novel core structure-based neural network. Detailed descriptions of this core structure and the overall architecture of the neural network are provided in the subsequent section.

\section{Core Structure Based Neural Network} 
%\textcolor{red}{The core structure-based neural network architecture is designed to be low-complexity, scalable, token-free, memory-efficient, and hardware-friendly. A neural network with these features can significantly facilitate the implementation of complex diffusion models on-device.}

%The core structure-based neural network consists of three main components: an initial module for image embedding extraction, core structure-based reusable blocks, and a final decoder module.

%\subsection{Core Structure}  
The core structure (transformer block) consists of multi-head attention followed by an MLP, with layer normalization applied prior to both components. The block diagram of the transformer block is illustrated in Figure~\ref{fig:core_structure}. 
%This core structure is utilized to build a hardware-friendly neural network architecture for modeling the reverse diffusion process. 
It is designed with fixed input and output dimensions. By stacking these core structures, we construct the overall neural network. Since the core structures have consistent input and output sizes, resource allocation remains straightforward and uniform.

\subsection{Proposed Neural Network}
We propose a neural network architecture using core structure for the diffusion process, incorporating a noise prediction network $\epsilon_{\theta}(x_t,t)$ (or equivalently, a score matching network $s_{\theta}(x_t,t)$), parameterized by Equation~\ref{eq:ddpm_loss} (or Equation~\ref{eq:ncsn_2}). This network estimates the noise to be removed during the reverse denoising process, based on the transition distribution from $x_t$. In the case of text-conditional image synthesis, the model also incorporates text conditioning $c$ as input, in addition to time $t$. Our proposed architecture is inspired by the pyramidal feature extraction process characteristic of CNNs. Initially, it extracts significant image features, and as the layers deepen, it identifies more detailed features. Initial convolutional blocks extract features, which are subsequently fed to fixed-size core structure. Finally, the features of the fixed sized core structure are given to decoder block match the image resolution of our choice. 
%\textcolor{red}{We experiment and report efficacy of the proposed neural network for conditional and unconditional image generation using diffusion models.}
The implementation details are discussed in Section~\ref{sec:implement}.

\subsection{Implementation details}
\label{sec:implement}
We demonstrate the efficacy of the core structure based proposed neural network designed for diffusion model using conditional (text-to-image) and unconditional (noise-to-image) image generation task. %\textcolor{red}{For conditional image generation we have mainly three inputs to give to the neural network, namely input noise, conditioning input and time embedding information. Where as for unconditional image generation we have inputs as noise and time embedding only.} 
Through experimentation, we observe that concatenating the time embedding before and after the initial convolution block yields the same effect. For text-conditional image synthesis, the CLIP~\cite{radford2021learning} text embedding is concatenated along 
the channel or embedding dimension of the initial convolution block output. The complete architecture is illustrated in Figure~\ref{fig:core_structure}.  

We design the core structure based neural network architecture by exploring various configurations at different sections. Through experimental analysis on CIFAR-10, we assess the impact of these design choices on key elements of the core structure, evaluating performance every 50K iterations. We report FID~\cite{heusel2017gans} on 10K samples to demonstrate effectiveness, alongside computational cost measured in GMAC.

\begin{comment}
\begin{figure*}
    \centering
    \includegraphics[width=1.0\linewidth]{figures/T2I_mscoco_core_lite.pdf}
  
     \caption{ Conditional diffusion model with repetative core structures.}
     \label{fig:core_structure}
\end{figure*}
\end{comment}

\begin{figure*}
\begin{subfigure}{0.45\linewidth}
  \centering
   \includegraphics[width=1.0\linewidth]{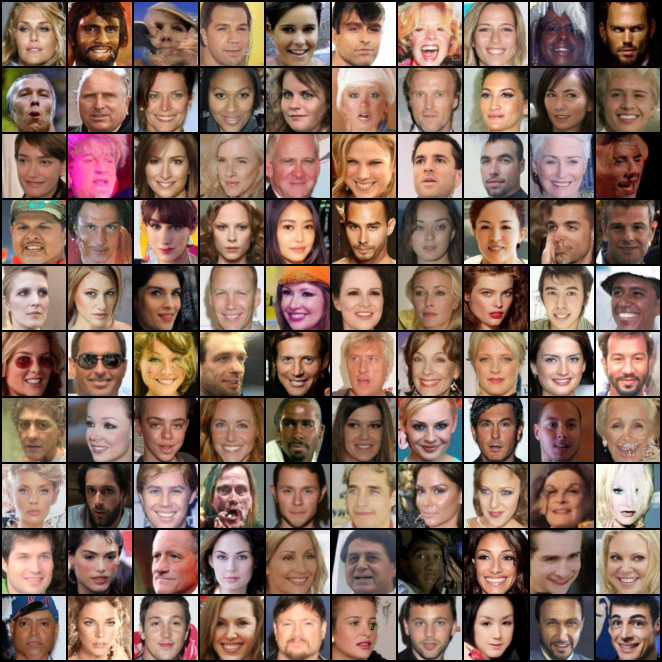}

   \caption{}
   \label{fig:uncond_celeba}
\end{subfigure}
\hfill
\begin{subfigure}{0.45\linewidth}
  \centering
   \includegraphics[width=1.0\linewidth]{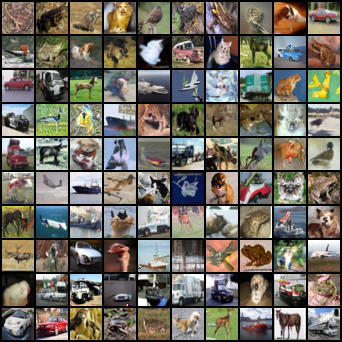}

   \caption{}
   \label{fig:uncond_cifar10}
\end{subfigure}
\caption{Unselected unconditional image samples from (a) CelebA and (b) CIFAR-10.}
\end{figure*}

\textbf{Initial convolution:}   
We employ two variants to obtain the initial embedding from the input image using an initial convolution block: one with stride $\mathcal{S}=2$ and the other with $\mathcal{S}=1$. Given any 2D input $\mathcal{X}_i$ where $\mathcal{X}_i \in \mathbb{R}^{H_i \times W_i \times C_i}$ with any batch size $B$, and 
applying any 2D convolutional layer $f(.)$ which takes input with channel size $C_i$ and outputs a 2D result with channel size $C_o$, denoted as $\mathcal{Y}_i = f(\mathcal{X}_i)$ such that $\mathcal{Y}_i \in \mathbb{R}^{H_o \times W_o \times C_o}$, and given a kernel 
size $(\mathcal{K}\times \mathcal{K})$, stride $\mathcal{S}$, and padding $P$, the output height $H_o$ and width $W_o$ are calculated as follows:

$ H_o = \lfloor \frac{H_i+2P-\mathcal{K}-1}{\mathcal{S}} +1 \rfloor \> , \>W_o = \lfloor \frac{W_i+2P-\mathcal{K}-1}{\mathcal{S}} +1 \rfloor $,
where $\lfloor \cdot \rfloor$ denotes the floor function, which returns the greatest integer less than or equal to its argument. Hence, with $\{\mathcal{S}=1, \mathcal{K} =3, P=1\}$, the output dimension $H_o = H_i$ and $W_o = W_i$. With $\{\mathcal{S}=2, \mathcal{K} =2, P=0\}$, the output dimension $H_o = \frac{H_i}{2}$ and $W_o = \frac{W_i}{2}$. 

As illustrated in Figure~\ref{fig:nonli_ini}, incorporating non-linearity in the initial convolution block results in improved image quality, evidenced by a reduction in the FID score. Additionally, using batch normalization degrades the process; therefore, we omit batch normalization, as shown in Figure~\ref{fig:norm_ini}. It is important to note the GMAC associated with stride $\mathcal{S}=2$, as depicted in the Figure~\ref{fig:gflops_s1_s2}. As we increase the model size by changing the number of core structures and the embedding dimension used, the computational cost remains low, with GMAC staying below 30. Additionally, the FID score decreases significantly as the parameter count increases, while the corresponding increase in GMAC remains minimal. 
%In contrast, ViT-based models use the same kernel size and stride for tokenization, which are equivalent to the patch dimensions. This requires synchronization of all patches, achieved by using positional embeddings at a later stage. The output is subsequently concatenated with the context embedding and then passed into the core structures.
The initial convolution block offers several advantages, including the efficient preservation of essential features, reduced computational complexity, and seamless compatibility with the target dimensions of the core structure.

\begin{comment}
\begin{figure}[t]
  \centering
   \includegraphics[width=1.0\linewidth]{figures/bubble_plt.pdf}

   \caption{GFlops with different embedding dimension and different number of blocks for $\mathcal{S}=2$.}
   \label{fig:gflops_s1_s2}
\end{figure}

\begin{figure}[t]
  \centering
   \includegraphics[width=1.0\linewidth]{figures/fid_plt_nonli_ini.pdf}

   \caption{Non linearity in initial embedding layer}
   \label{fig:nonli_ini}
\end{figure}

\begin{figure}[t]
  \centering
   \includegraphics[width=1.0\linewidth]{figures/fid_plt_norm_ini.pdf}

   \caption{Normalization in initial embedding layer}
   \label{fig:norm_ini}
\end{figure}
\end{comment}

\textbf{Time embedding:} 
The time embedding block generates a 1D tensor for each index of the batch element. The time embedding is obtained by passing through a linear layer and subsequently reshaped to concatenate with the image embedding. This concatenation can occur either before or after the initial convolution block. If concatenated before initial convolution block, the dimension of the embedding is chosen such that it can be reshaped to $\mathbb{R}^{H_i \times W_i}$. If concatenated after the initial convolution block, 
it should be reshaped to $\mathbb{R}^{H_o \times W_o}$. In both cases, the concatenation occurs along the channel or embedding dimension. If the concatenation is performed before the initial embedding, 
the input channel of the initial convolution block is set as the sum of the input image channels plus one. 
An important observation is that both variants produce a similar effect on the generated image quality.

\textbf{Context embedding:}
The context embedding is passed through a linear layer and then reshaped to $\mathbb{R}^{H_o \times W_o}$ and concatenated along the channel dimension of initial convolution block output features. This increases the channel dimension from the embedding dimension 
to the embedding dimension plus the context embedding dimension, which is 77 for the CLIP text embedding in text-conditional image generation. This increased channel dimension is then reshaped 
back to the embedding dimension using an additional linear layer after concatenating the context embedding.

\textbf{Core structure configaration:}
We conducted experiments by concatenating $N$ core structures, with $N = {12, 24, 32}$ core blocks. Increasing the number of such blocks (making the model deeper) and increasing the embedding dimension $L$ (making the model wider) improves model performance in terms of the FID score, as shown in Figure~\ref{fig:mod_si}. However, this improvement comes with an increase in the complexity, as shown in Figure~\ref{fig:gflops_s1_s2}.

\begin{figure*}
  \centering
  \begin{tabular}{c}
   \subfloat[]{\includegraphics[width=1.0\linewidth]{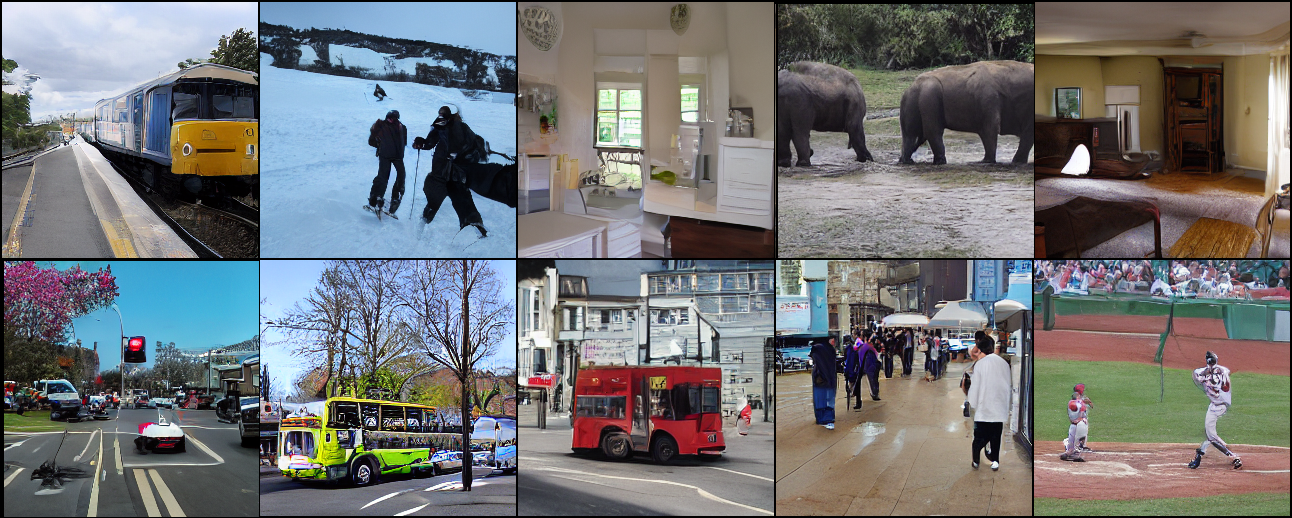} \label{fig:text_cond_syn_stoic}} \\
   \subfloat[]{\includegraphics[width=1.0\linewidth]{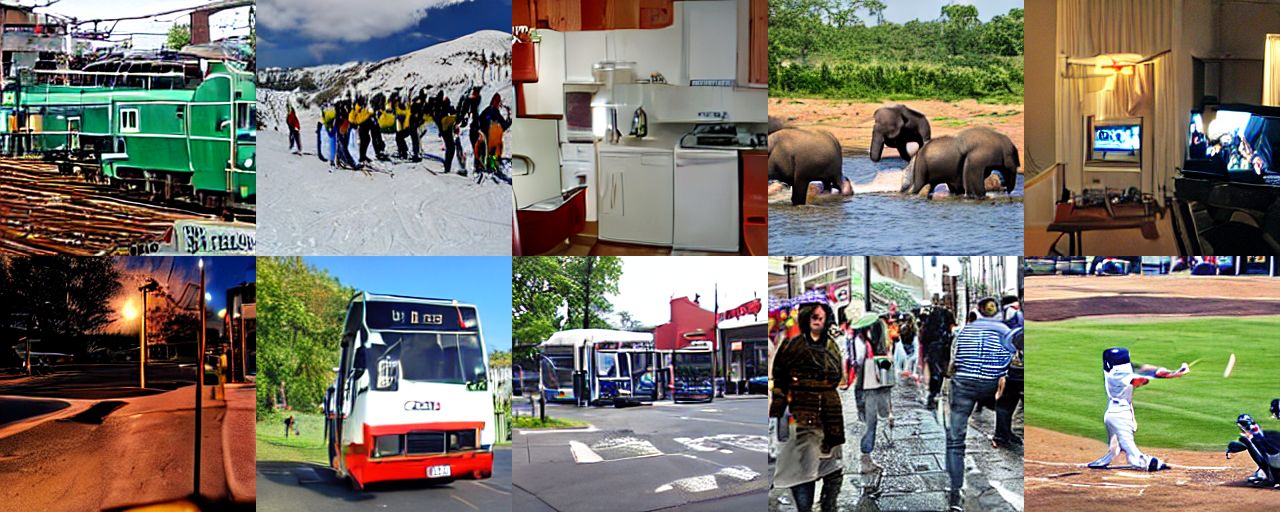} \label{fig:text_cond_syn_uvit}}
   
    \end{tabular}
   \caption{Qualitative comparison between (a) STOIC and (b) U-ViT for text-conditional image generation on the MS-COCO dataset. STOIC-$\mathcal{S}-2$ maintains a comparable parameter count ($\sim$43M) to U-ViT, while achieving significantly lower computational complexity, with approximately 11 GMACs compared to U-ViT's 43 GMACs. Text prompts: 1. A train is coming down the tracks. 2. A group of skiers are preparing to ski down a mountain. 3. A small kitchen with a low ceiling. 
   4. A group of elephants walking in muddy water. 5. A living area with a television and a table. 6. A road with traffic lights, street lights and cars. 7. A bus driving in a city area with traffic signs. 
   8. A bus pulls over to the curb close to an intersection. 9. A group of people are walking and one is holding an umbrella. 10. A baseball player taking a swing at an incoming ball.
   }
   \label{fig:text_cond_syn}
%\vspace{-0.2cm}
\end{figure*}

\textbf{Decoder Block:}
The sequential output of the core structure is finally fed to the decoder block to generate an output matching the size of the input image of our choice. Initially, this block includes an additional layer normalization step followed by a linear layer to reduce the embedding dimension $\mathbb{R}^{(H_o \times W_o) \times L} \rightarrow \mathbb{R}^{(H_o \times W_o) \times C}$, where $C$ represents the input channel dimension and $L$ denotes the embedding dimension of the core structures. We utilized two variants to scale it down: a linear layer or simply slicing the initial $C$ embedding tensors, depending on the stride $\mathcal{S}$ used in the initial embedding block and the input channel length $C$. It has been observed that the slicing operation slightly improves performance, as shown in Figure~\ref{fig:lin_sli}. Subsequently, the tensor of size $\mathbb{R}^{(H_o \times W_o) \times C}$ is reshaped to $\mathbb{R}^{H_o\times W_o \times C}$ before being fed into the final 2D convolutional layer. This layer then maps the tensor from $\mathbb{R}^{H_o\times W_o \times C}$ to the input dimension of $\mathbb{R}^{H_i\times W_i \times C}$. While the final convolutional layer enhances model performance, we observed that both Conv2D and ConvTranspose2D 
produce the same FID score, as shown in Figure~\ref{fig:conv_conv_t}.

\begin{comment}
\begin{figure}[t]
  \centering
   \includegraphics[width=1.0\linewidth]{figures/fid_plt_lin_sli.pdf}

   \caption{Linear layer vs slicing operation in the Decoder block.}
   \label{fig:lin_sli}
\end{figure}

\begin{figure}[t]
  \centering
   \includegraphics[width=1.0\linewidth]{figures/fid_plt_conv_convt.pdf}

   \caption{ Conv2D vs ConvTranspose2D operation in the decoder block}
   \label{fig:conv_conv_t}
\end{figure}
\end{comment}

\section{Experiments}
We evaluate the performance of our proposed neural network architecture, \textbf{S}calable, \textbf{TO}ken-free \textbf{I}nitial \textbf{C}onvolution (\textbf{STOIC}), for diffusion models, utilizing two variants: STOIC-$\mathcal{S}$-1 (with stride $\mathcal{S}=1$) and STOIC-$\mathcal{S}$-2 (with stride $\mathcal{S}=2$). These variants are assessed across both unconditional and text-conditional image generation tasks.

\subsection{Experimental setup}
\textbf{Dataset:}
For unconditional image synthesis, we use the CIFAR-10~\cite{krizhevsky2009learning} and CelebA~\cite{liu2015deep} datasets. CIFAR-10 consists images of resolution $32 \times 32$ and total training samples are 50K, while CelebA comprises images of resolution $64 \times 64$ and 162,770 training samples. For text-conditional image synthesis, we employ the MSCOCO~\cite{lin2014microsoft} dataset, which includes $256 \times 256$ resolution images, with 82,783 training samples, each accompanied by 5 textual descriptions.

\textbf{Training:} We train the CIFAR-10 dataset with a batch size of 128 for 500K iterations. For the MSCOCO dataset, training is conducted for 1M iterations with a batch size of 256. The CelebA dataset is trained with a batch size of 64 for 500k. We use the AdamW optimizer~\cite{loshchilov2017decoupled}, with a learning rate set at $1\times10^{-4}$. During inference, images are sampled from the trained DDPM model using classifier-free guidance~\cite{ho2022classifier}. We utilize the Euler-Maruyama SDE sampler~\cite{song2020score} for unconditional image synthesis and the DPM Solver ODE sampler~\cite{lu2022dpm}, following~\cite{bao2023all} for text conditional image synthesis. We employ the Fr\'echet Inception Distance (FID)~\cite{heusel2017gans}, the standard metric for evaluating generative models. 

\begin{table*}
\scalebox{0.9}{
\begin{subtable}{0.40\textwidth}
  \centering
  \begin{tabular}{c c c}
    \toprule
    \multicolumn{3}{c}{\textbf{CIFAR-10}}\\
    \hline
    \textbf{Method} & \textbf{FID-50K} $\downarrow$ & \textbf{Parameters} $\downarrow$\\
    \midrule
    \multicolumn{3}{c}{\textbf{Model based on UNet}} \\
    \hline
    DDPM~\cite{ho2020denoising}&  3.17 & 36M\\
    IDDPM~\cite{nichol2021improved} & 2.90 & 53M\\
    EDM~\cite{karras2022elucidating} & \textbf{1.97} & 56M\\
    VDM~\cite{kingma2021variational} & 7.41 & -\\
    NCSN++~\cite{song2020score} & 2.20 & 62M\\
    PNDM~\cite{liu2022pseudo} & 3.26 & 62M\\
    Scoreflow~\cite{song2021maximum} & 3.98 & -\\
    \hline
    \multicolumn{3}{c}{\textbf{Model based on ViT}}\\
    \hline
    Gen-ViT~\cite{yang2022your} & 20.2 & 11M\\
    Styleformer~\cite{park2022styleformer} &2.82& -\\
    UViT~\cite{bao2023all}& 3.11 & 44M\\
    \hline
    STOIC-$\mathcal{S}$-2 & 3.5 & 82M\\
    \cellcolor{green!25}STOIC-$\mathcal{S}$-1 & \cellcolor{green!25}3.05 & \cellcolor{green!25}82M\\
    \bottomrule
  \end{tabular}
  \caption{}
  \label{tab:cifar_comp}
\end{subtable}
}
\hfill
\scalebox{0.9}{
\begin{subtable}{0.55\textwidth}
  \centering
  \begin{tabular}{c c c c}
    \toprule
    \multicolumn{4}{c}{\textbf{CelebA 64$\times$64}}\\
    \hline
    \textbf{Method} &\textbf{Type}& \textbf{FID-50K} $\downarrow$ & \textbf{Parameters} $\downarrow$\\
    \midrule
    TransGAN~\cite{jiang2021transgan}&GAN& 12.23&-\\
    HIDCGAN~\cite{curto2017high}&GAN&8.77&-\\
    \hline
    \multicolumn{4}{c}{\textbf{Model based on UNet}}\\
    \hline
    DDIM~\cite{song2020denoising} &Diffusion& 3.26 & 79M\\
    Soft Truncation~\cite{kim2021soft} &Diffusion& 1.92 & 62M\\
    NCSN++~\cite{song2020score} &Diffusion& 3.95 & 62M\\
    PNDM~\cite{liu2022pseudo} &Diffusion& 2.71 & 62M\\
    INDM~\cite{kim2022maximum} &Diffusion& 1.75 & 118M\\
    DDGM~\cite{nachmani2021non} &Diffusion& 2.92 & -\\
    \hline
    \multicolumn{4}{c}{\textbf{Model based on ViT}}\\
    \hline
    Styleformer~\cite{park2022styleformer}&Diffusion& 3.66& -\\
    UViT~\cite{bao2023all} &Diffusion& 2.87 & 44M\\
    \hline
    STOIC-$\mathcal{S}$-2 &Diffusion& 3.6 & 88M\\
    \cellcolor{green!25}STOIC-$\mathcal{S}$-1 &Diffusion \cellcolor{green!25}& \cellcolor{green!25}\textbf{1.6} & \cellcolor{green!25}88M\\
    \bottomrule
  \end{tabular}
  \caption{}
  \label{tab:celeba_comp}
\end{subtable}
}
\caption{(a) A comparison of FID scores on the unconditional CIFAR-10 dataset across various diffusion-based models. (b) A comparison of FID scores on the unconditional CelebA dataset across different GAN-based and diffusion-based models.}
\end{table*}

\begin{figure*}
\begin{subfigure}{0.32\linewidth}
  \centering
   \includegraphics[width=1.0\linewidth]{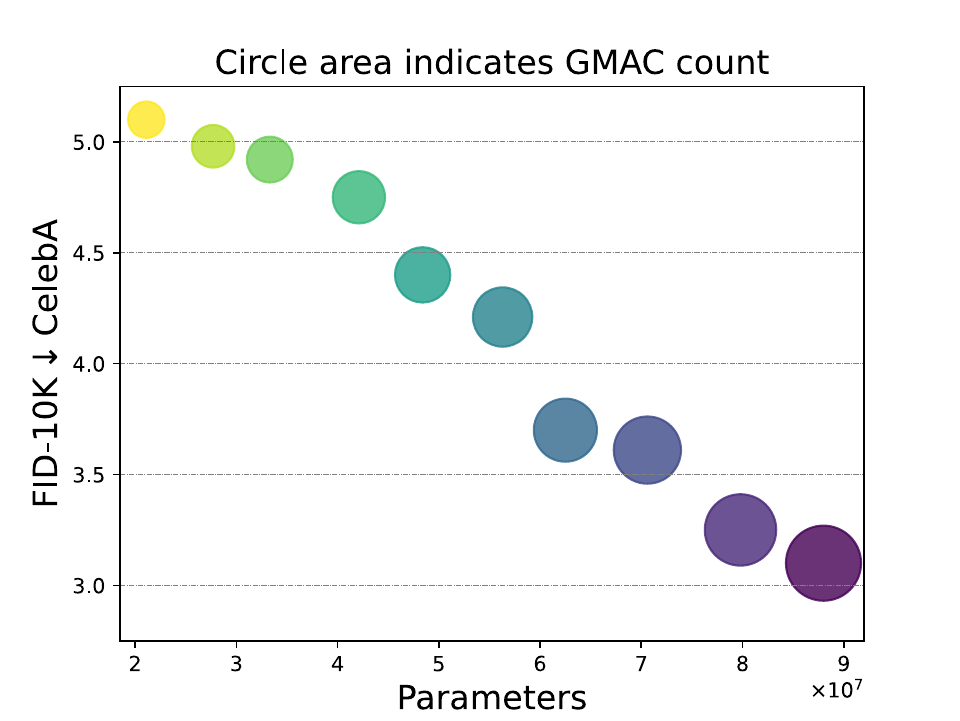}
   %\caption{Scaling paremeters with $\{\mathcal{S}=1, \mathcal{K} =3, P=1\}$ Con2D at the initial embedding layer. GFlops size is multiplied with 15 to show for better visibility}
   \caption{}
   \label{fig:param_fid_1}
\end{subfigure}
\hfill
\begin{subfigure}{0.32\linewidth}
  \centering
   \includegraphics[width=1.0\linewidth]{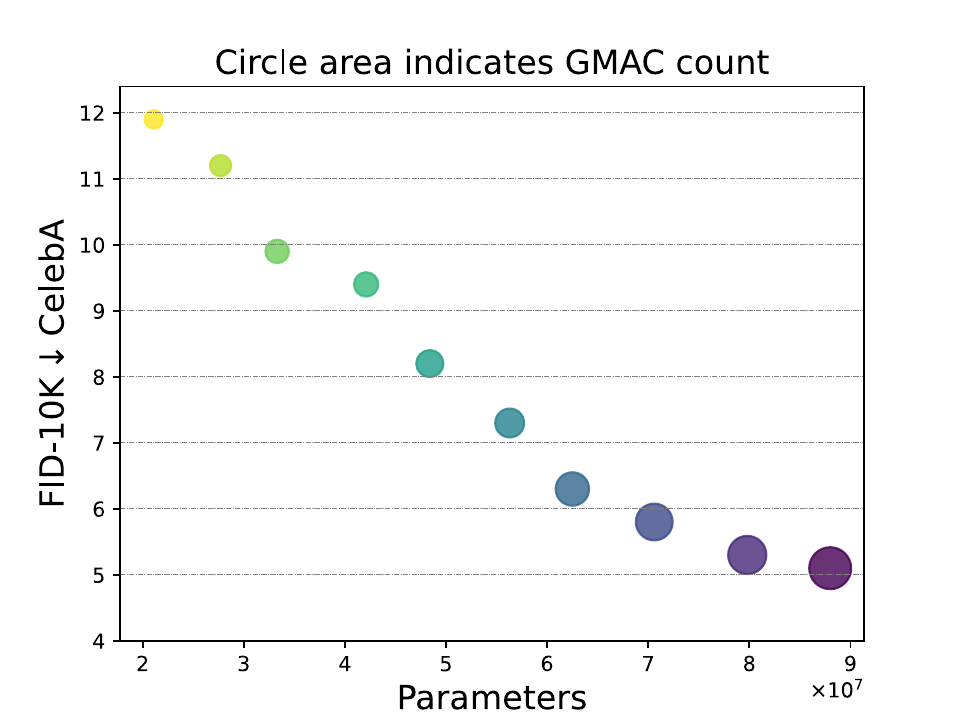}
   %\caption{ Scaling paremeters with $\{\mathcal{S}=2, \mathcal{K} =2, P=0\}$ Conv2D at the initial embedding layer. GFlops size is multiplied with 15 to show for better visibility.}
   \caption{}
   \label{fig:param_fid_2}
\end{subfigure}
\hfill
\begin{subfigure}{0.32\linewidth}
  \centering
   \includegraphics[width=1.0\linewidth]{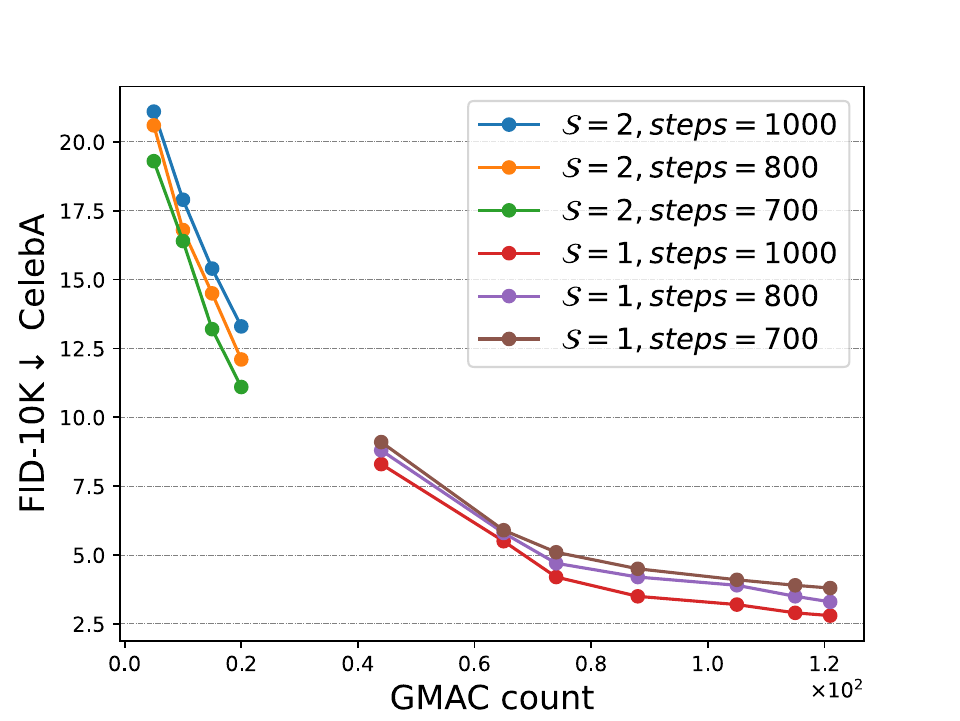}
   %\caption{ For the variants $\{\mathcal{S}=2$ and $\{\mathcal{S}=1$ the models trained for 400K iterations, we compute FID-10K using different sampling steps.}
   \caption{}
   \label{fig:fid_plt_flops_sampling}
\end{subfigure}
   \caption{(a) Scaling parameters with $\{\mathcal{S}=1, \mathcal{K} =3, P=1\}$ Conv2D at the initial embedding layer. (b) Scaling parameters with $\{\mathcal{S}=2, \mathcal{K} =2, P=0\}$ Conv2D at the initial embedding layer. In both (a) and (b), FID is reported after completing 500K iterations, with GMAC multiplied by 15 for enhanced visibility. It is evident that the FID reduction is nearly linear in both cases, demonstrating the scalability of the STOIC architecture. (c) For the variants STOIC-$\mathcal{S}$-1 and STOIC-$\mathcal{S}$-2 the models trained for 500K iterations, we compute FID-10K using different sampling steps.}
\end{figure*}

\begin{table*}
  \centering
  \scalebox{0.9}{
  \begin{tabular}{c c c c c}
    \toprule
    \textbf{Model} & \textbf{Type} & \textbf{FID-30K $\downarrow$} & \textbf{Parameters $\downarrow$} & \textbf{Datasets} \\
    \midrule 
    \multicolumn{5}{c}{Text conditional image synthesis on external large dataset} \\
    \hline
    DALL-E~\cite{ramesh2021zero} & Autoregressive &  28 & 12B & DALL-E Dataset \\
    CogView~\cite{ding2021cogview}& Autoregressive & 27.1 & 4B & Internal Dataset \\
    LAFITE~\cite{zhou2111lafite}& GAN & 26.94 & 75M+151M{$^\dagger$} & Conceptual Captions 3M\\
    GLIDE~\cite{nichol2021glide}& Diffusion & 12.24 & 3.5B+1.5B{$^{\dagger\dagger}$} & DALL-E Dataset \\
    DALL-E 2~\cite{ramesh2022hierarchical}& Autoregressive, Diffusion & 10.39 & 4.5B+700M{$^{\dagger\dagger}$} & DALL-E Dataset \\
    Imagen~\cite{saharia2022photorealistic}& Diffusion & 7.27 & 2B & Internal Dataset, LAION \\
    Parti~\cite{yu2022scaling}& Autoregressive & 7.23 & 20B &LAION \\
    \hline
    \multicolumn{5}{c}{Text conditional image synthesis on MSCOCO} \\
    \hline
    AttnGAN~\cite{xu2018attngan}& Autoregressive & 35.49 &  230M & MSCOCO \\
    DM-GAN~\cite{zhu2019dm}& GAN & 32.64 & 46M & MSCOCO \\
    VQ-Diffusion~\cite{gu2022vector}& Diffusion & 19.75 & 370M & MSCOCO\\
    XMC-GAN~\cite{zhang2021cross}& GAN & 9.33 & 166M & MSCOCO \\
    Frido~\cite{fan2023frido}& Diffusion & 8.97 & 512M+186M{$^\dagger$}+68M{$^\ddagger$} & MSCOCO \\
    LAFITE~\cite{zhou2111lafite}& GAN & 8.12 & 75M+151M{$^\dagger$} & MSCOCO\\
    \hline
    UViT~\cite{bao2023all}& Latent Diffusion & \textbf{5.48} & 58M(28{$^{\ddagger\ddagger}$}) +123M{$^\dagger$}+84M{$^\ddagger$} & MSCOCO \\
    LDM~\cite{rombach2022high}& Latent Diffusion & 12.63 & 1.45B+123M{$^\dagger$}+84M{$^\ddagger$} & MSCOCO \\
    %\cellcolor{green!25}STOIC-$\mathcal{S}$-1 & \cellcolor{green!25} Latent Diffusion & \cellcolor{green!25} -- & \cellcolor{green!25} 101M+123M(TE)+84M(AE) & \cellcolor{green!25} MSCOCO \\
    \cellcolor{green!25}STOIC-$\mathcal{S}$-2 & \cellcolor{green!25}Latent Diffusion & \cellcolor{green!25} 8.69 & \cellcolor{green!25}101M(22.35{$^{\ddagger\ddagger}$})+123M{$^\dagger$}+84M{$^\ddagger$} & \cellcolor{green!25}MSCOCO \\
    \bottomrule
  \end{tabular}
  }
  \caption{Comparison of text-conditional image synthesis across various methods and datasets. Supplementary components, when reported in the respective papers, are also included. Here, {$^{\dagger}$} indicates the use of text encoders such as CLIP, {$^{\ddagger}$} refers to autoencoders, {$^{\dagger\dagger}$} represents the use of a super-resolution module, and {$^{\ddagger\ddagger}$} denotes the GMAC count.}
  \label{tab:table_our_results}
\end{table*}

\begin{comment}
\begin{figure*}
  \centering
   \includegraphics[width=1.0\linewidth]{figures/compare_uvit.pdf}

   \caption{Text conditional image generation with different prompts by comparison with U-ViT\cite{bao2023all}}
   \label{fig:text_cond_syn_comp}
\end{figure*}
\end{comment}

\subsection{Main result}
We compare our results with various Vision Transformer-based diffusion models as well as several U-Net-based diffusion models for both unconditional and text-conditional image synthesis tasks.

\textbf{Unconditional image synthesis:}
We compared our unconditional image synthesis on the CIFAR-10 and CelebA datasets. For Vision Transformer-based models, we compared with GenViT~\cite{yang2022your}, Styleformer~\cite{park2022styleformer} and UViT~\cite{bao2023all}, the latter being a ViT that employs long skip connections outside the transformer blocks and incorporates time after the initial embedding layer.

As shown in Table~\ref{tab:cifar_comp}, STOIC performs comparably to UNet and Vision Transformer models on unconditional CIFAR-10, while significantly outperforming GenViT~\cite{yang2022your}. Specifically, STOIC-$\mathcal{S}$-2, with 82M parameters, achieved an FID score of 3.5. When utilizing the same parameter count of 82M, but with the STOIC-$\mathcal{S}$-1, the FID improved to 3.05. This result surpasses UViT's~\cite{bao2023all} FID of 3.11 (achieved with 44M parameters), PNDM's~\cite{liu2022pseudo} FID of 3.26, and is comparable to IDDPM's~\cite{nichol2021improved} FID of 2.9 (which employs a UNet architecture with 53M parameters) and Styleformer's~\cite{park2022styleformer} FID of 2.82. The samples generated for CIFAR-10 are illustrated in Figure~\ref{fig:uncond_cifar10}.

On the CelebA dataset, STOIC demonstrates superior performance, surpassing the existing baselines, as shown in Table~\ref{tab:celeba_comp}. Our two variants, STOIC-$\mathcal{S}$-1 and STOIC-$\mathcal{S}$-2, both with 88M parameters, achieved FID scores of 1.6 and 3.6, respectively. STOIC-$\mathcal{S}$-1 outperforms UViT's FID of 2.87 with 44M parameters, DDIM's FID of 3.26 with 79M parameters, and INDM's FID of 1.75 with 118M parameters. This result also exceeds Soft Truncation's~\cite{kim2021soft} FID of 1.9, which softens the smallest diffusion time and estimates it using a random variable. Additionally, STOIC surpasses NCSN++'s~\cite{song2020score} FID of 3.95, a variant of NCSN that utilizes SDE for both forward and reverse diffusion processes. We experimented with three different values for the number of sampling steps using the Euler-Maruyama sampler—700, 800, and 1,000—resulting in FID-10K scores of 3.8, 3.3, and 2.8, respectively, as illustrated in Figure~\ref{fig:fid_plt_flops_sampling}. The samples generated for CIFAR-10 are illustrated in Figure~\ref{fig:uncond_celeba}.

\begin{comment}
\begin{figure}
  \centering
   \includegraphics[width=1.0\linewidth]{figures/celeba_unconditional.pdf}

   \caption{Unconditional image synthesis on CelebA.}
   \label{fig:uncond_celeba}
\end{figure}

\begin{figure}
  \centering
  \fbox{\rule{0pt}{2in} \rule{0.9\linewidth}{0pt}}
   %\includegraphics[width=1.0\linewidth]{figures/unconditional_celeba.pdf}

   \caption{Unconditional image synthesis on CIFAR10.}
   \label{fig:uncond_cifar10}
\end{figure}

\begin{figure*}
  \centering
   \includegraphics[width=1.0\linewidth]{figures/text_conditional_mscoco.png}

   \caption{Text conditional image generation. Text prompts: 1. A green train is coming down the tracks. 2. A group of skiers are preparing to ski down a mountain. 3. A small kitchen with a low ceiling. 
   4. A group of elephants walking in muddy water. 5. A living area with a television and a table. 6. A road with traffic lights, street lights and cars. 7. A bus driving in a city area with traffic signs. 
   8. A bus pulls over to the curb close to an intersection. 9. A group of people are walking and one is holding an umbrella. 10. A baseball player taking a swing at an incoming ball.
   }
   \label{fig:text_cond_syn}
\end{figure*}
\end{comment}

\textbf{Text conditional image synthesis:}
We evaluated the performance of our STOIC-based diffusion model for text-conditional image synthesis on the MSCOCO dataset, comparing it against various benchmarks, as shown in Table~\ref{tab:table_our_results}.
%We demonstrated the text conditional image generation using LDM models on MSCOCO dataset. 
The text description of the MSCOCO dataset is encoded using the text encoder of CLIP~\cite{radford2021learning}. This embedding is subsequently fed to our model for text-conditional image synthesis. We employed a pretrained autoencoder from LDM~\cite{rombach2022high} to encode images of size $256 \times 256 \times 3$ into latent representations of size $32 \times 32 \times 4$ for the MSCOCO dataset. We conducted experiments with two variants,  STOIC-$\mathcal{S}$-1 and STOIC-$\mathcal{S}$-2, and with different numbers of model parameters. Notably, STOIC-$\mathcal{S}$-2 achieves an FID of 8.69 with 101M parameters, outperforming GAN-based models such as VQ-Diffusion (370M parameters) and XMC-GAN (166M parameters), despite having fewer parameters. 
%Furthermore, as shown in Table~\ref{tab:table_our_results}, STOIC-$\mathcal{S}$-1 achieves an improved FID of -- with a parameter count of 101M, outperforming other methods that do not access large external datasets during the training of generative models.
%%%%%%%%%%%%%%%%%%%%%%%%%%%%%%%%%%%%%%%%%%%%%%%%%%%%%%%%%%%%%%%%%%%%%%
In Figure~\ref{fig:text_cond_syn}, we present the results of our text-to-image synthesis model across various text prompts with UViT~\cite{bao2023all}. 
%Additionally, in Figure~\ref{fig:text_cond_syn_comp}, we provide a comparative analysis between the outputs generated by UViT~\cite{bao2023all} and our model, using the same random seed to ensure a fair comparison. The UViT model utilizes 44 million parameters and operates with 44 GMAC, while our model has a parameter count of 28 million and requires 5.34 GMAC.
We observed that the STOIC-based diffusion model, suitable for deployment on mobile devices for inference, generates higher-quality samples with improved alignment between the text and the generated images. In contrast, the UViT model requires higher GMAC (with our GMAC count being 22.35 compared to UViT's 28) for comparable performance and incurs high latency on hardware devices due to long skip connections.
%For example, given the `` a couple of horses standing next to each other on a field'' text a baseball player swinging a bat at a ball’’, U-Net generates neither the bat nor the ball.

%In contrast, our STOIC generates the ball with even a smaller number of parameters, and our STOIC further generates the bat. We hypothesize this is because texts and images interact at every layer in our STOIC, which is more frequent than U-Net that only interact at cross attention layer. 

\begin{comment}
\begin{figure}[t]
  \centering
   \includegraphics[width=1.0\linewidth]{figures/param_fid_1.pdf}

   \caption{Scaling paremeters with $\{\mathcal{S}=1, \mathcal{K} =3, P=1\}$ Con2D at the initial embedding layer. GFlops size is multiplied with 15 to show for better visibility}
   \label{fig:param_fid_1}
\end{figure}

\begin{figure}[t]
  \centering
   \includegraphics[width=1.0\linewidth]{figures/param_fid_2.pdf}

   \caption{ Scaling paremeters with $\{\mathcal{S}=2, \mathcal{K} =2, P=0\}$ Conv2D at the initial embedding layer. GFlops size is multiplied with 15 to show for better visibility}
   \label{fig:param_fid_2}
\end{figure}
\end{comment}

\begin{comment}
\begin{figure*}
  \centering
   \includegraphics[width=1.0\linewidth]{figures/compare_uvit.pdf}

   \caption{Text conditional image generation. Comparison with U-ViT}
   \label{fig:text_cond_syn}
\end{figure*}
\end{comment}

\textbf{Scaling parameters and GMAC:}
We present ablation studies examining how increasing the parameters of STOIC-$\mathcal{S}$-1 and STOIC-$\mathcal{S}$-2 by altering the embedding dimension and the number of core structures affects GMAC, parameter count, and the corresponding reduction in FID score.

\underline{STOIC-$\mathcal{S}$-1:} \textit{Better FID with low parameters.}
With stride $\mathcal{S}=1$ we see even with very low parameter count (high GMAC) the model is generating good FID score as shown in Figure~\ref{fig:param_fid_1}. This is because
even though the parameter count is low initial convolution is processing the full image embedding and is generating meaningful output which is of the same size. This shows better performance than STOIC-$\mathcal{S}$-2 in terms of FID score.

\underline{STOIC-$\mathcal{S}$-2:} \textit{light in terms of GMAC to get good FID}  
With stride $\mathcal{S}=2$, we observe that it is possible to increase parameter count, while maintaining a relatively low GMAC, compared to $\mathcal{S}=1$, to obtain a state-of-the-art FID score, as illustrated in Figure~\ref{fig:param_fid_2}. This indicates that good results can be achieved even with low GMAC.

\section{Conclusion}
Our study shows that the STOIC model offers a robust framework for unconditional and conditional image synthesis, outperforming benchmarks with more efficient use of computational resources. We present an architecture closely resembling the Vision Transformer, offering the flexibility of scalable transformer blocks and reduced complexity, but without the overhead of tokenization or the need for positional embeddings. In text-conditional image generation, the diffusion process in latent space highlights the model's ability to balance performance and resource efficiency, while the unconditional image generation results further demonstrate its effectiveness in pixel space.

{\small
\bibliographystyle{ieee_fullname}
\bibliography{egbib}
}
%\newpage
%\newpage

\begin{comment}

\clearpage

\appendix

\section{Additional Samples}

\begin{figure*}
  \centering
   \includegraphics[width=0.9\linewidth]{figures/additional_samples_celeba.eps}

   \caption{Unconditional image generation on CelebA. }
   \label{fig:text_cond_syn}
\end{figure*}
\end{comment}

%{
%    \small
%    \bibliographystyle{ieeenat_fullname}
%    \bibliography{main}
%}

\end{document}